\documentclass[conference]{IEEEtran}  
\usepackage{graphicx}
\usepackage{tabularx,booktabs}
\usepackage[usestackEOL]{stackengine}
\usepackage{adjustbox}
\usepackage{blindtext}

\makeatletter
\newcommand{\linebreakand}{%
  \end{@IEEEauthorhalign}
  \hfill\mbox{}\par
  \mbox{}\hfill\begin{@IEEEauthorhalign}
}

\begin{document}

\makeatother
\title{\LARGE \bf
AGE RANGE ESTIMATION USING MTCNN AND VGG-FACE MODEL
}


  

\author{
    \IEEEauthorblockN{Dipesh Gyawali\textsuperscript1, Prashanga Pokharel\textsuperscript2, Ashutosh Chauhan\textsuperscript3, Subodh Chandra Shakya\textsuperscript4}
    \IEEEauthorblockA{InfoDevelopers Pvt. Ltd. \\ Lalitpur, Nepal
     \\}
     \IEEEauthorblockN{(\textsuperscript1dipesh9393, \textsuperscript2pokharel997,
     \textsuperscript3Cashutosh6711,
     \textsuperscript4subshakya591)@gmail.com
    \\}}



\maketitle
\thispagestyle{empty}
\pagestyle{empty}

\begin{abstract}

The Convolutional Neural Network has amazed us with its usage on several applications. Age range estimation using CNN is emerging due to its application in myriad of areas which makes it a state-of-the-art area for research and improve the estimation accuracy. A deep CNN model is used for identification of people's age range in our proposed work. At first, we extracted only face images from image dataset using MTCNN to remove unnecessary features other than face from the image. Secondly, we used random crop technique for data augmentation to improve the model performance. We have used the concept of transfer learning in our research. A pretrained face recognition model i.e  VGG-Face is used to build our model for identification of age range whose performance is evaluated on Adience Benchmark for confirming the efficacy of our work. The performance in test set outperformed existing state-of-the-art by substantial margins.

\end{abstract}
\medskip


\section{{INTRODUCTION}}

The face images are used for different applications recently in face recognition[1], surveillance system[15],  emotion identification[16], and smart attendance for multifarious purposes. Identifying the age range using CNN has become a challenging task as the age range among the adults seems to be similar than the child and aged people[8]. The facial features among the child and old people vary drastically from the adult people which makes it easier to identify the age of child and old people[7]. But the identification among the adult age becomes cumbersome due to the presence of almost the same facial features.
\par

By far, CNN has become most effective for processing image data in object detection, face recognition, and image classification. Image data consists of a lot of parameters which are difficult for a basic neural network to process. A lot of matrix calculation needs to be handled for 2D image data where CNN is a most effective neural network as different convolutional layers in CNN are used to extract the feature map of images and the convolutional process with the filter make it easy to process image data.
\par

In our research, we have identified the age range among various ages of people using novel approach. Transfer learning has been considered more effective than building our own custom CNN model for a limited number of dataset. The model which was pre-trained for image classification and face recognition[3] is used for estimating the age range.

 \begin{figure}
  \centering
  \includegraphics[scale=0.30]{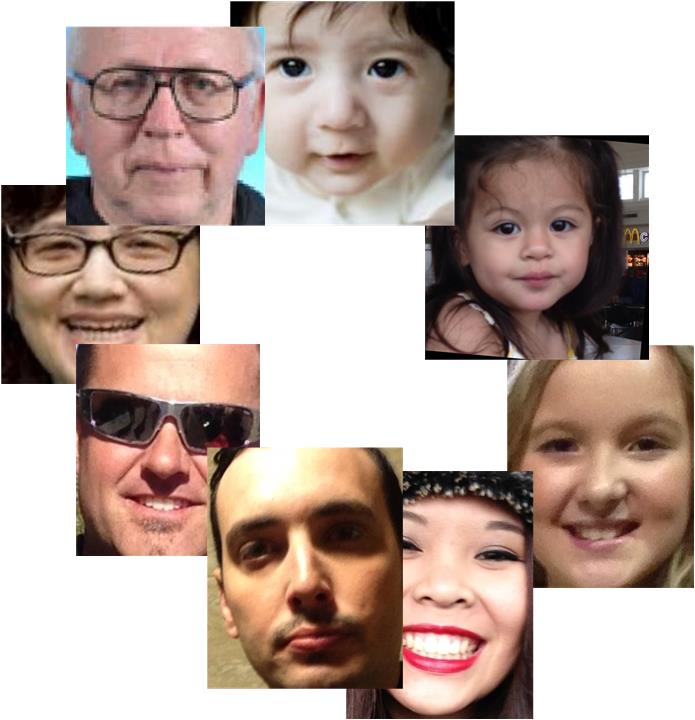}
  \caption{Face images of different age groups on Adience Benchmark[11]}
\end{figure}

\medskip

\section{{RELATED WORKS}}
A dataset including one million faces of celebrities are prepared which eventually improved face recognition accuracy[1]. Face recognition using new dataset across various identities minimizing the label noise[2],using either a single image or set of faces[3], analysis and survey[4] is performed. Age Estimation is done using VGG-Face model[5], the ranking SVM[6] and finding sufficient embedding space by applying manifold learning methods which models data with multiple linear regression function[7] effectively. A craniofacial growth model is proposed which models growth related variations in shape of human faces[8]. Different faces are divided into several minute regions for the extraction of Local Binary Pattern(LBP) to use as a face descriptor by concatenating into feature vectors[9]. Two new approaches, Ranking-CNN[10] based on rank relationship and deep CNN architecture for low quality face images[24] are proposed for identifying age from different perspectives. Using unfiltered faces on Adience Benchmark[11] and in the presence of limited learning data using deep CNN[12], identification of gender and age is performed. Various research challenges in addition with recent survey are delineated for research in face recognition along with age estimation[13]. The effects of aging on performance of age invariant face recognition are surveyed[14] and analyzed using deep features analysis[21].
\par

The above works showed that a handful of research tasks have already been carried out using different metrics and methods. Although a plethora of works have been done, a significant improvement in the identification of age range across adult age group due to similar facial features had not been achieved. In this experiment, we have initially extracted only the facial features from joint face alignment using MTCNN network and later identified age range using VGG-Face model which significantly reduced overfitting and improved identification among adult age group as compared to previous works. Also, a comparative analysis has been done across multiple face models to analyze efficiency of our work using unconstrained face images.

\medskip

\section{{DATA PREPARATION}}
MTCNN (Multi-task Cascaded Convolutional Networks) is a network[17] which is used in our proposed work to extract the facial features. MTCNN contains 3 CNN steps, each step called as Proposal Network (P-net), Refine Network (R-net) and Output Network (O-net). All the input images are fed to Proposal Network (P-Net), which is a CNN model. The candidate window inside the image with their bounding box regression vector is received as output from P-net. We create an image pyramid, in order to detect faces of all different sizes as shown in Fig. 2.
\medskip
 \begin{figure}[hbt!]
  \centering
  \includegraphics[scale=0.18]{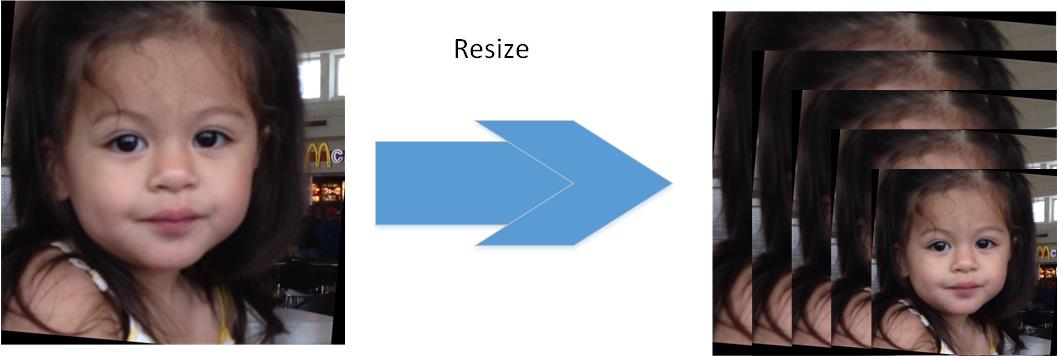}

  \includegraphics[scale=0.18]{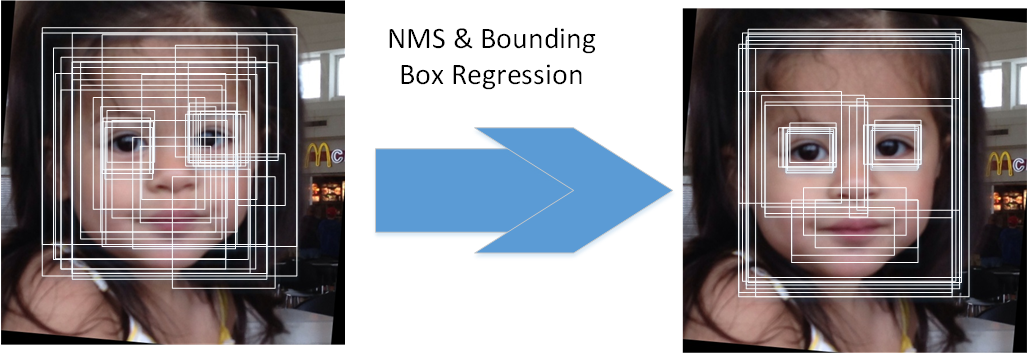}
  \caption{Image pyramids and NMS and bounding box regression (P-Net)}
  \end{figure}

After feeding those candidates to Refined Network (R-Net), we have identified regression vector of bounding box and used Non-Max Suppression (NMS) to find and integrate densely overlapped candidates as shown in Fig. 3. The outcomes from this network are passed to O-Net.
\medskip

 \begin{figure}[hbt!]
  \centering
  \includegraphics[scale=0.18]{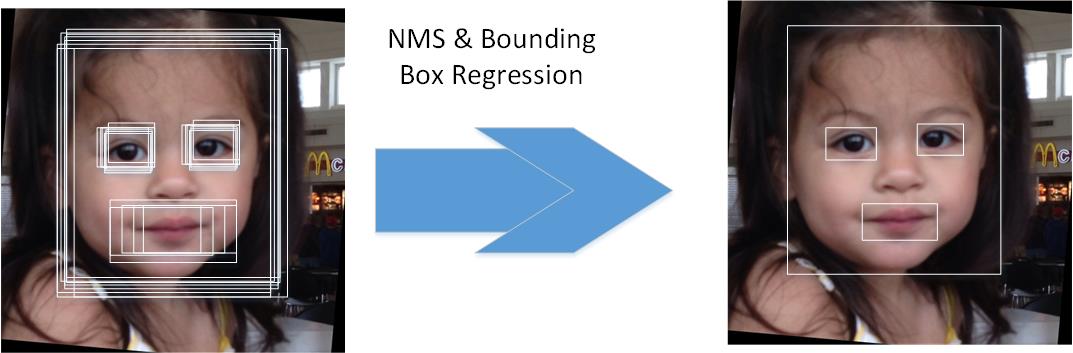}
  \caption{NMS and bounding box regression (R-Net)}
  \end{figure}

Finally, Output Network (O-Net) provides three output, coordinate of bounding box, the coordinate of five facial landmarks, and the confidence level of each box as shown in Fig. 4. This bounding box image is saved as a new image which is then passed into the proposed model. 
\medskip

 \begin{figure}[hbt!]
  \centering
  \includegraphics[scale=0.18]{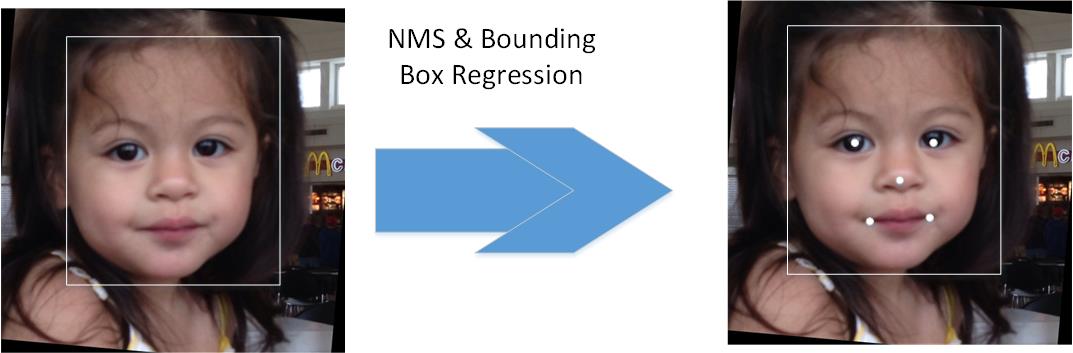}
  \caption{Output-network}
  \end{figure}

Images in the dataset are rescaled to 256x256 pixels. These images are extracted of size 224x224 pixels using random crop[18]. Horizontal flip and rotation is applied to the extracted images for data augmentation purpose. All age groups were divided into 8 classes. For each class, images were splitted 80-20 for training and validation. Finally this dataset is fed to the network.

\section{{PROPOSED SYSTEM}}
A large number of images are required for building our own custom model to ameliorate its performance. In order to prevent overfitting, we have used transfer learning for building our CNN network. The model is well validated and tested after optimizing it using appropriate hyperparameters. The VGG-Face model performs well on Adience Benchmark[11] due to its less number of CNN layers which can be useful for small dataset. 

\subsection{{Architecture}}

For the purpose of age estimation, the pre-trained model used for face recognition task[3] is considered extremely useful. The age estimation is generally performed by observing the facial features of different age groups[7]. Due of this, for the estimation of age robustly, model previously trained on face recognition can be used. By the use of pre-trained models, there will be less chance of overfitting. Also, we have used MTCNN for extracting the facial features while training in our dataset.

 \begin{figure}[hbt!]
  \centering
  \includegraphics[scale=0.19]{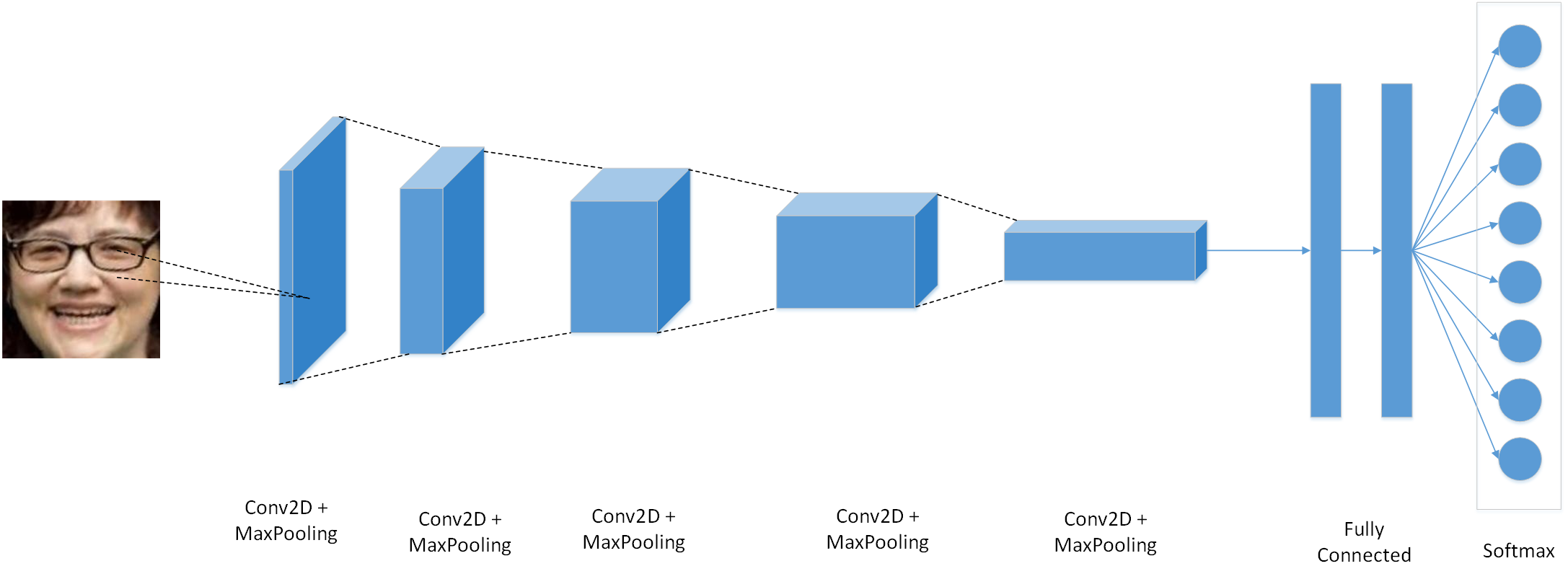}
  \caption{CNN Architecture}
  \end{figure}

\medskip
To identify a better age range, we have used pre-trained model VGG-Face. The model performed pretty well using VGG-Face due to its less number of CNN layers and appropriate tuning of hyperparameters. Eight convolutional layers comprising of Conv2D \& Pooling and three fully connected layers makes up a overall of eleven layers of VGG-Face model. All the layers except fully connected softmax layers were used for training purposes. The layers from VGG-Face are kept frozen first and then additional layers are added to the VGG-Face model as per our requirements. Two fully connected layers are added in corresponding with one dropout layer and one softmax output layer to identify the age range for 8 classes. The layers and the respective dimensions used in our network are represented as in the Table I.

\begin{table*}
\centering

\caption{Our proposed CNN architecture using VGG-Face Model}
\begin{adjustbox}{width=1\textwidth}
\begin{tabular}{cccccccccccccc} 
\hline
\begin{tabular}[c]{@{}c@{}} Layer\\type\\name \end{tabular} & \begin{tabular}[c]{@{}c@{}}0\\input\\- \end{tabular} & \begin{tabular}[c]{@{}c@{}}1\\conv\\conv1\_1~\end{tabular} & \begin{tabular}[c]{@{}c@{}}2\\relu\\relu1\_1~\end{tabular} & \begin{tabular}[c]{@{}c@{}}3\\conv\\conv1\_2~\end{tabular} & \begin{tabular}[c]{@{}c@{}}4\\relu\\relu1\_2~\end{tabular} & \begin{tabular}[c]{@{}c@{}}5\\mpool\\pool1 \end{tabular} & \begin{tabular}[c]{@{}c@{}}6\\conv\\conv2\_1~\end{tabular} & \begin{tabular}[c]{@{}c@{}}7\\relu\\relu2\_1 \end{tabular} & \begin{tabular}[c]{@{}c@{}}8\\conv\\conv2\_2~\end{tabular} & \begin{tabular}[c]{@{}c@{}}9\\relu\\relu2\_2~\end{tabular} & \begin{tabular}[c]{@{}c@{}}10\\mpool\\pool2 \end{tabular} & \begin{tabular}[c]{@{}c@{}}11\\conv\\conv3\_1~\end{tabular} & \begin{tabular}[c]{@{}c@{}}12\\relu\\relu3\_1 \end{tabular} \\ 
\hline
support & - & 3 & 1 & 3 & 1 & 2 & 3 & 1 & 3 & 1 & 2 & 3 & 1 \\
filt dim & - & 3 & - & 64 & - & - & 64 & - & 128 & - & - & 128 & - \\
num filts & - & 64 & - & 64 & - & - & 128 & - & 128 & - & - & 256 & - \\
stride & - & 1 & 1 & 1 & 1 & 2 & 1 & 1 & 1 & 1 & 2 & 1 & 1 \\
pad & - & 1 & 0 & 1 & 0 & 0 & 1 & 0 & 1 & 0 & 0 & 1 & 0 \\
\hline

\small
\begin{tabular}[c]{@{}c@{}} Layer\\ type\\ name \end{tabular} & 
\begin{tabular}[c]{@{}c@{}}13\\conv\\conv3\_2~\end{tabular} &
\begin{tabular}[c]{@{}c@{}}14\\ input\\ relu3\_ 2\end{tabular} &
\begin{tabular}[c]{@{}c@{}}15\\ conv\\ conv3\_3~\end{tabular} & \begin{tabular}[c]{@{}c@{}}16\\ relu\\ relu3\_3~\end{tabular} & \begin{tabular}[c]{@{}c@{}}17\\ mpool\\ pool3 \end{tabular} & \begin{tabular}[c]{@{}c@{}}18\\ conv\\ conv4\_1~\end{tabular} & \begin{tabular}[c]{@{}c@{}}19\\ relu\\ relu4\_ 1\end{tabular} & \begin{tabular}[c]{@{}c@{}}20\\ conv\\ conv4\_ 2\end{tabular} & \begin{tabular}[c]{@{}c@{}}21\\ relu\\ relu4\_2~\end{tabular} & \begin{tabular}[c]{@{}c@{}}22\\ conv\\ conv4\_ 3\end{tabular} & \begin{tabular}[c]{@{}c@{}}23\\ relu\\ relu\_3~\end{tabular} & \begin{tabular}[c]{@{}c@{}}24\\ mpool\\ pool4 \end{tabular} & \begin{tabular}[c]{@{}c@{}}25\\conv\\conv5\_1~\end{tabular} \\ 
\hline
support & 3 & 1 & 3 & 1 & 2 & 3 & 1 & 3 & 1 & 3 & 1 & 2 & 3 \\
filt dim & 256 & - & 256 & - & - & 256 & - & 512 & - & 512 & - & - & 512 \\
num filts & 256 & - & 256 & - & - & 512 & - & 512 & - & 512 & - & - & 512 \\
stride & 1 & 1 & 1 & 1 & 2 & 1 & 1 & 1 & 1 & 1 & 1 & 2 & 1 \\
pad & 1 & 0 & 1 & 0 & 0 & 1 & 0 & 1 & 0 & 1 & 0 & 0 & 1 \\
\hline


\begin{tabular}[c]{@{}c@{}} Layer\\ type\\ name \end{tabular} &
\begin{tabular}[c]{@{}c@{}}26\\relu\\relu5\_1~\end{tabular} &
\begin{tabular}[c]{@{}c@{}}27\\ conv\\ conv5\_2~\end{tabular} & \begin{tabular}[c]{@{}c@{}}28\\ relu\\ relu5\_2~\end{tabular} & \begin{tabular}[c]{@{}c@{}}29\\conv\\conv5\_3~~\end{tabular} & \begin{tabular}[c]{@{}c@{}}30\\relu\\ conv5\_3~\end{tabular} & \begin{tabular}[c]{@{}c@{}}31\\mpool\\pool5~\end{tabular} & \begin{tabular}[c]{@{}c@{}}32\\ conv\\fc6~\end{tabular} & \begin{tabular}[c]{@{}c@{}}33\\ relu\\ relu6~\end{tabular} & \begin{tabular}[c]{@{}c@{}}34\\dropout\\dropout7~\end{tabular} & \begin{tabular}[c]{@{}c@{}}35\\conv\\fc8~\end{tabular} & \begin{tabular}[c]{@{}c@{}}36\\relu\\relu7~\end{tabular} & \begin{tabular}[c]{@{}c@{}}37\\softmax\\prob\end{tabular} & ~ ~ ~ ~~\\ 
\cline{1-13}
support & 1 & 3 & 1 & 3 & 1 & 2 & 7 & 1 & 1 & 1 & 1 & 1 & \\
filt dim & - & 512 & - & 512 & - & - & 512 & - & - & 1000 & - & - & \\
num filts & - & 512 & - & 512 & - & - & 1000 & - & - & 100 & - & - & \\
stride & 1 & 1 & 1 & 1 & 1 & 2 & 1 & 1 & 1 & 1 & 1 & 1 & \\
pad & 0 & 1 & 0 & 1 & 0 & 0 & 0 & 0 & 0 & 0 & 0 & 0 & \\
\cline{1-13}
\end{tabular}
\end{adjustbox}
\end{table*}
\medskip

\subsection{{Loss Function}}
We have multiple classes of age-range ranging from 0-2 to 60-100 with a total of eight classes. The actual and predicted probability distributions for all eight classes are identified whose average differences are summarized with a score calculated using cross entropy[19]. A total of eight nodes are present at the output layer for each age range in which 'softmax' activation is used to predict the probability across each classes. Due to a multiclass classification problem and effectiveness, we have used categorical cross entropy as a loss function[19].
\medskip
\begin{equation}
CE = -\sum_{c=1}^{M} t_{o,c}log\left ( P_{o,c} \right )
\end{equation}

\medskip
where t is either 0 or 1 if label class c is the correct classification for observation o. M is a total number of classes. P is predicted probability of class c. 
\medskip

\subsection{{Optimizer}}
Due to efficient computation, little memory requirements and presence of intuitive interpretations of hyperparameters, we have used Adam optimizer[20] in our research which is considered more effective comparatively with other optimizers. 

\subsection{{Training}}
We used transfer learning(VGG-Face) to train our model. The convolutional layer parameters of VGG-Face are not changed and kept frozen. We optimize the fully connected layers parameters by adding additional two fully connected layers, one dropout layer and a final output layer to the VGG-Face model. Number of filters in the first of two fully connected layers are 1000 and 100 consecutively. In these fully connected layers we use ‘relu’ activation function. We use 0.3 (30\% chance of setting a neuron’s output value to zero) values for one dropout layer. 'Softmax' activation is used in the final layer to classify the age range. ‘Adam optimizer’ is used as an optimizer with ‘categorical cross entropy’ as a loss function in our model. 

\subsection{{Prediction}}
After training the model, we have used the Adience Benchmark to evaluate the model performance. A test image is rescaled to 256x256 pixels. Then five images are extracted of size 224x224 pixels. The four images are obtained from four corners of the image and the final image is obtained from the center of the original test image as shown in Fig. 6. These five images are fed into our trained neural network to calculate the softmax probability output vector of each of the five images. These output scores vector of these five images were averaged. Average result helps to reduce the impact of poor quality and low-resolution images.

 \begin{figure}[hbt!]
  \centering
  \includegraphics[scale=0.10]{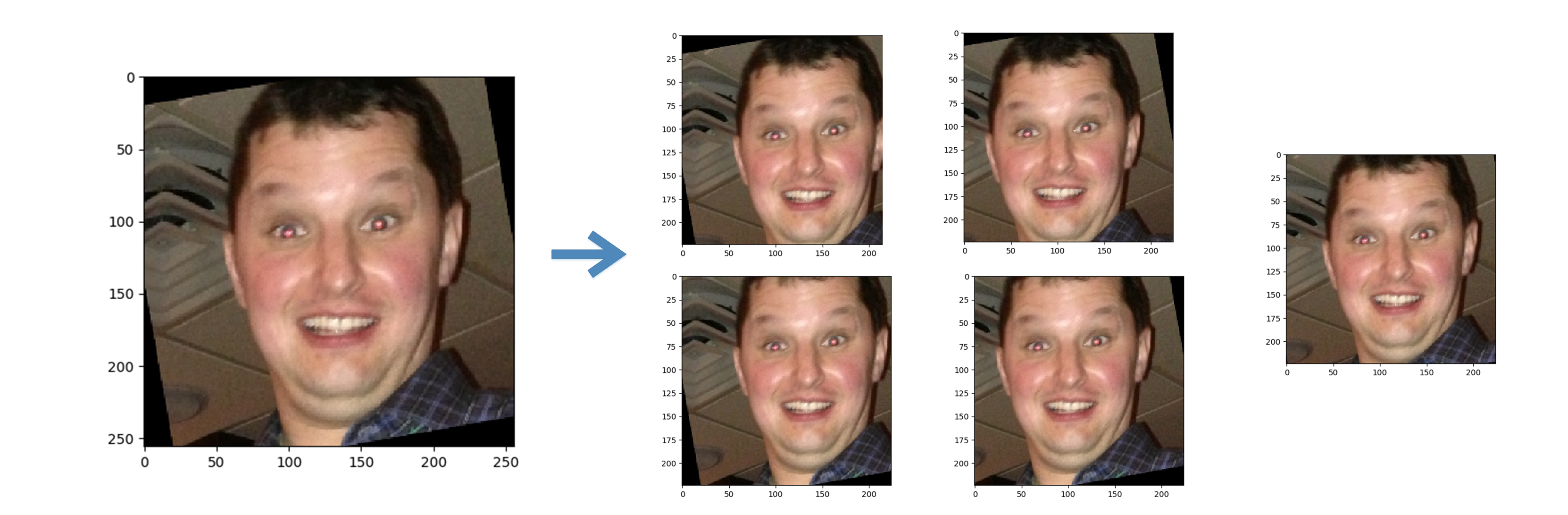}
  \caption{Image Rescaling to 256x256 pixels initially. Five images are extracted from each four corners and centre which are resized to 224x224 for feeding into the network.}
  \end{figure}

\section{{EXPERIMENTS AND RESULTS}}

A wide known and flexible deep learning library i.e Keras consisting of Tensorflow as a backend engine is used in our proposed work for effectiveness. Training was done on NVIDIA GTX 2080 GPU with 4352 CUDA cores. Our training time took approximately four hours. We trained VGG-Face Model for 50 epochs with a batch size of 64. The model performance is evaluated using Adience benchmark. The Adience benchmark is a large-scale face dataset which consists of 19487 multiple face images with 8 classes of age range covering large variation in facial expression, pose, occlusion, resolution and illumination. The exact accuracy on Adience benchmark came out to be 70.96\%. We have used this Adience benchmark for appraising the efficacy of our work. Different age range's label used in our research with the corresponding number of images per label on Adience Benchmark are shown in Table II. 

\begin{table}
\centering
\caption{The Adience Benchmark}
\begin{adjustbox}{width={(\textwidth/2)-10mm},totalheight={\textheight},keepaspectratio}%
\begin{tabular}{cccccccccc} 
\hline
 & 0-2 & 4-6 & 8-13 & 15-20 & 25-32 & 38-43 & 48-53 & 60+ & Total \\ 
\hline
Male & 745 & 928 & 934 & 734 & 2308 & 1294 & 392 & 442 & 8192 \\
Female & 682 & 1234 & 1360 & 919 & 2589 & 1056 & 433 & 427 & 9411 \\ 
\hline
Both & 1427 & 2162 & 2294 & 1653 & 4897 & 2350 & 825 & 869 & 19487 \\
\hline
\end{tabular}
\end{adjustbox}
\end{table}

Consequently, our proposed model outperformed the previous methods reported in [5], [11], and [12].The accuracy was compared with the previous work done for age estimation in which our model performs better as compared to other performances. Table III shows exact accuracy and 1-off accuracy from the previous work and our proposed work using VGG-Face architecture. Our proposed work outperforms the previous work and confirm the efficiency of our proposed work. The use of MTCNN approach has significantly improved accuracy on Adience Benchmark. Fig. 7 shows the confusion matrix on Adience Benchmark.

\begin{table}
\centering
\caption{Age estimation result on the Adience Benchmark}
\begin{adjustbox}{width={(\textwidth/2)-10mm},totalheight={\textheight},keepaspectratio}
\begin{tabular}{lll} 
\hline
Method & Exact Accuracy & 1-off Accuracy \\ 
\hline
{[}11] & 45.1 & 79.5 \\
{[}12] using single crop & 49.5 & 84.6 \\
{[}12] using over-sample & 50.7 & 84.7 \\
{[}5] & 59.9 & 90.57 \\ 
\hline
Proposed work & \textbf{70.96}  & \textbf{92.7}  \\
\hline
\end{tabular}
\end{adjustbox}
\end{table}

 \begin{figure}[hbt!]
  \centering
  \includegraphics[scale=0.25]{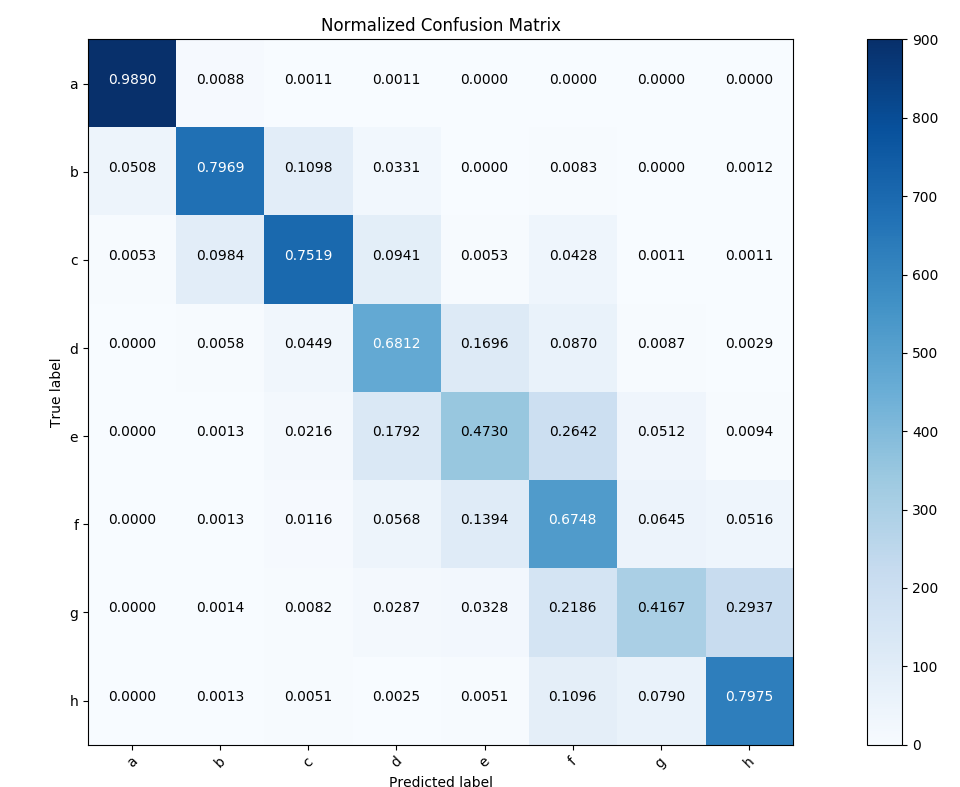}
  \caption{\textbf{Normalized Confusion Matrix for Adience Benchmark.} The label is marked from a-h to make ease while building confusion matrix. The label is defined as a(0-2),b(4-6),c(8-13),d(15-20),e(25-32),f(38-43),g(48-53),h(60+).}
  \end{figure}

The true label and the predicted label are denoted from indices of each rows and each columns respectively in the confusion matrix. The occurences of prediction is given by number showing on each cell. In the leading diagonal of confusion matrix, the true label and predicted label are equal and other off-diagonal elements represent occurence mislabeled by classifier. There are more correct predictions if there are higher values in the leading diagonal. The confusion matrix in Fig. 7. clearly shows probability of prediction. Our result from the confusion matrix depicts that the highest accuracy is 98.90\% which of (0-2) age range as shown in Fig. 7. The highest accuracy is due to the distinctive features which enable the classifier to distinguish easily. The images of middle age groups consist of almost similar features which resulted in less accuracy as compared to (0-2) age range and (60+) age range. In our work, the age group of (48-53) and (25-32) is highly misclassified with  41.67\% and 47.30\%. The classification report on Adience Benchmark using VGG-Face model is shown in the Table IV.

\begin{table}
\centering
\caption{Classification Report on Adience Benchmark}
\begin{adjustbox}{width={(\textwidth/2)-10mm},totalheight={\textheight},keepaspectratio}
\begin{tabular}{ccccc} 
\hline
\begin{tabular}[c]{@{}c@{}} Age Range From\\Adience Dataset \end{tabular} & Precision & Recall & F1-Score & Support \\ 
\hline
0-2 & 0.95 & 0.99 & 0.97 & 1427 \\
4-6 & 0.86 & 0.80 & 0.83 & 2162 \\
8-13 & 0.81 & 0.75 & 0.78 & 2294 \\
15-20 & 0.60 & 0.68 & 0.64 & 1653 \\
25-32 & 0.58 & 0.47 & 0.52 & 4897 \\
38-43 & 0.49 & 0.67 & 0.57 & 2350 \\
48-53 & 0.66 & 0.42 & 0.51 & 825 \\
60+ & 0.70 & 0.80 & 0.75 & 869 \\ 
\hline
Accuracy & \multicolumn{1}{l}{} & \multicolumn{1}{l}{} & 0.71 & 19487 \\
macro avg & 0.71 & 0.70 & 0.69 & 19487 \\
weighted avg & 0.72 & 0.71 & 0.71 & 19487 \\
\hline
\end{tabular}
\end{adjustbox}
\end{table}

We have also used VGG-Face2 Model to analyze the efficacy of our task. Our proposed work using VGG-Face model proved to be far better than VGG-Face2 model which is shown in the Table V. There are more numbers of CNN layers on VGG-Face2 as compared to VGG-Face. As we had not enough face image data, the VGG-Face2 model was subjected to overfitting and could not perform well so that the performance across VGG-Face2 model seems very poor as compared to VGG-Face model.

 \begin{figure*}[hbt!]
  \centering
  \includegraphics[scale=0.45]{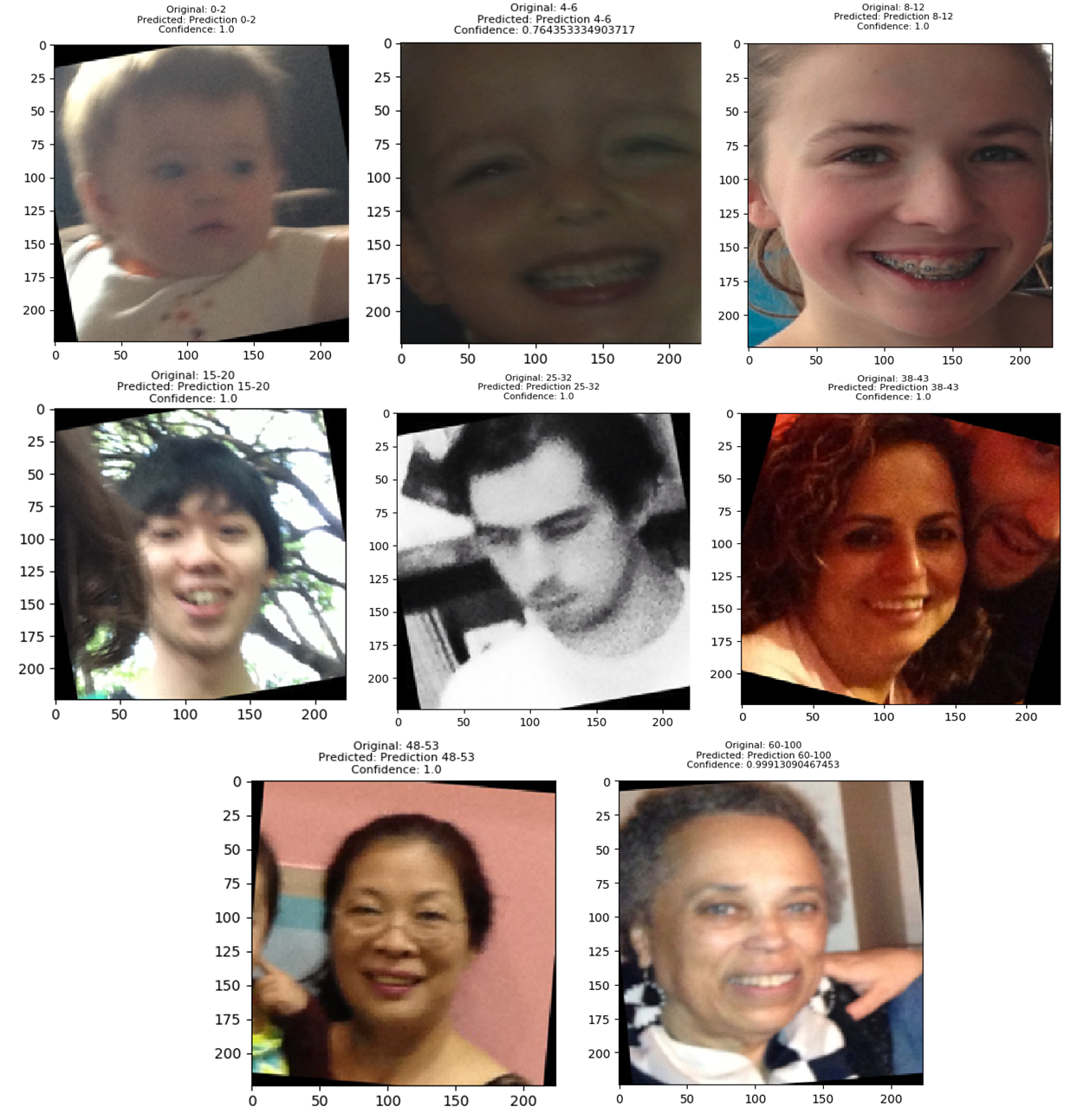}
  \caption{Some of the correctly classified images on Adience Benchmark}
  \end{figure*}
  
   \begin{figure*}[hbt!]
  \centering
  \includegraphics[scale=0.50]{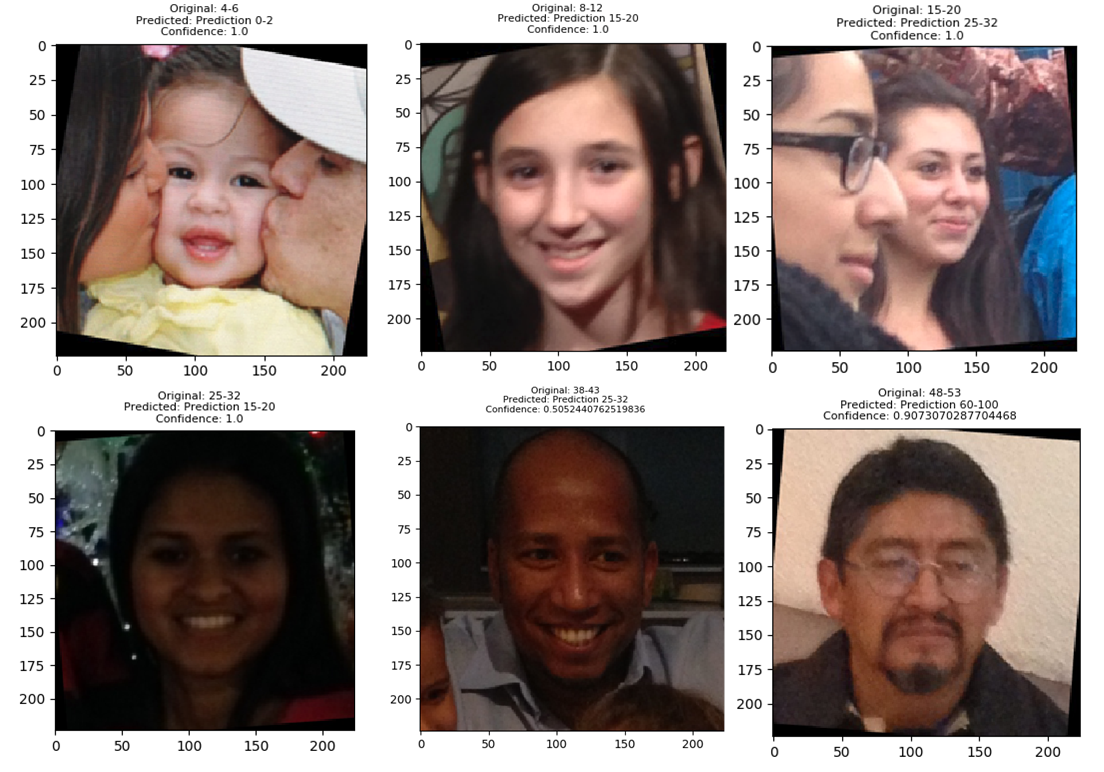}
  \caption{Some of the misclassified images on Adience Benchmark}
  \end{figure*}

Fig. 8 and Fig. 9 shows some of correctly classified and misclassified images on Adience Benchmark with their corresponding confidence level. The misclassified classes are mainly of adults due to same facial features. The age range of child and old age were almost classified correctly. The number of misclassified class could be improved further if we have high number of images for training. Different models can be used if there is copious amount of data in addition with proper pre-training process which might improve our proposed work.

\begin{table}
\centering
\caption{Exact Accuracy on Different Face Models}
\begin{adjustbox}{width={(\textwidth/2)-10mm},totalheight={\textheight},keepaspectratio}
\begin{tabular}{ccc} 
\hline
Label & \begin{tabular}[c]{@{}c@{}}Proposed model \\using VGG-Face\end{tabular} & \begin{tabular}[c]{@{}c@{}}Model using \\VGG-Face2\end{tabular} \\ 
\hline
0-2 & 98.90 & 80.40 \\
4-6 & 79.69 & 36.52 \\
8-13 & 75.19 & 29.64 \\
15-20 & 68.12 & 52.92 \\
25-32 & 47.30 & 10.37 \\
38-43 & 67.48 & 13.08 \\
48-53 & 41.67 & 49.89 \\
60+ & 79.75 & 77.86 \\ 
\hline
\textbf{Overall Exact Accuracy} & \textbf{70.96} & \textbf{42.02} \\
\hline
\end{tabular}
\end{adjustbox}
\end{table}

\section{{CONCLUSION}}

We have used the VGG-Face model and MTCNN to extract the facial features for estimating the age range for our research work. MTCNN helped to extract only the facial features from the image data which helps to determine the most germane features from the face. Our research work outperforms the previous work by almost 12\% on the Adience Benchmark. Due to the small number of dataset for training and large number of layers for feature extraction, the VGG-Face2 model is subjected to overfitting which does not perform well while evaluating its performance. Usage of MTCNN and fine tuning the VGG-Face model significantly improved the network’s performance for age range estimation. 
\par

Furthermore, there are still some misclassified images with high confidence because of the same facial features, lightness, occlusions and multiple person in an image. If the number of images in various conditions are increased and pre-training tasks of CNN network are performed significantly, the model performance can increase further. Also, VGG-Face2 model performance can be improved by using high number of face images and fine-tuning the network. The exact accuracy is still low due to the small number of dataset available for age estimation which could be improved in future works.


\medskip
\section*{{ACKNOWLEDGMENT}}
Our research is based upon work done at InfoDevelopers Pvt. Ltd. We would like to express gratitude towards our supervisor Mr. Sandesh Pandey for supervising and evaluating our research work. We are also thankful to Info Developers Private Limited for providing us the research platform and materials regarding our research procedures.


\end{document}